\PassOptionsToPackage{hyphens}{url}
\documentclass{article} 
\usepackage{iclr2026_conference,times}
\ifdefined\XeTeXversion
\usepackage{fontspec}
\fi

\usepackage{amsmath,amsfonts,bm}

\def\eqref#1{equation~\ref{#1}}

\def\1{\bm{1}}

\DeclareMathAlphabet{\mathsfit}{\encodingdefault}{\sfdefault}{m}{sl}
\SetMathAlphabet{\mathsfit}{bold}{\encodingdefault}{\sfdefault}{bx}{n}

\usepackage{hyperref}
\usepackage{url}
\usepackage{booktabs}
\usepackage{multirow}
\usepackage{xcolor}
\usepackage{graphicx}
\usepackage{array}
\usepackage{float}
\usepackage{placeins}
\usepackage{needspace}
\usepackage{fvextra}

\newcommand{\shortname}{RoboARCade}

\title{\shortname: Benchmarking Agentic\\Robot Control}

\author{Satvik Sharma, Hengyuan Hu, Mihika Kamat, Samrat Sahoo, Dorsa Sadigh, Jeannette Bohg \\
Stanford University
}

\definecolor{hhcolor}{HTML}{8F4A55}
\definecolor{arcorange}{HTML}{D9602B}

\iclrfinalcopy 
\begin{document}

\maketitle

\begin{abstract}


Improvements in multi-modal capabilities have enabled agents to act directly as robot policies, a setting we refer to as agentic robot control (ARC). We introduce \shortname{}, a benchmark of long-horizon manipulation tasks across three embodiments designed to test in-context adaptation, memory, and reasoning for precise control. The tasks build on existing open-source benchmarks, which we make harder by widening the range of initial conditions. RoboARCade is easy to use and extend: all tasks run on a single simulation backend, MuJoCo, each task is a self-contained data pack, and agents connect through interchangeable interfaces, so new tasks, agents, and interfaces can be added independently. The benchmark separates leading agents: GPT-6 Astra achieves 49.8\% compared with 43.6\% for Claude Fable 5.1, leaving substantial room for improvement. Comparing two interfaces, we find that a richer visual interface adds only 1.5 to 6.8 points at 1.7 to 2.3 times the cost. Failures stem mostly from choosing a poor manipulation strategy or executing a good one imprecisely, rarely from misunderstanding the scene, and agents incorrectly claim success on 24\% to 39\% of attempted subtasks. Agents can also improve without human input: one successful demonstration in context raises Astra’s score from 53.5\% to 73.0\% on a five-task subset, and a prompt the agent writes and revises itself reaches 72.0\%.
Code, videos and full traces are available at \href{https://roboarcade.org}{robo\textcolor{arcorange}{arc}ade.org}.
\end{abstract}

\section{Introduction}
\suppressfloats[t]

\begin{figure}[t]
\centering
\includegraphics[width=\linewidth]{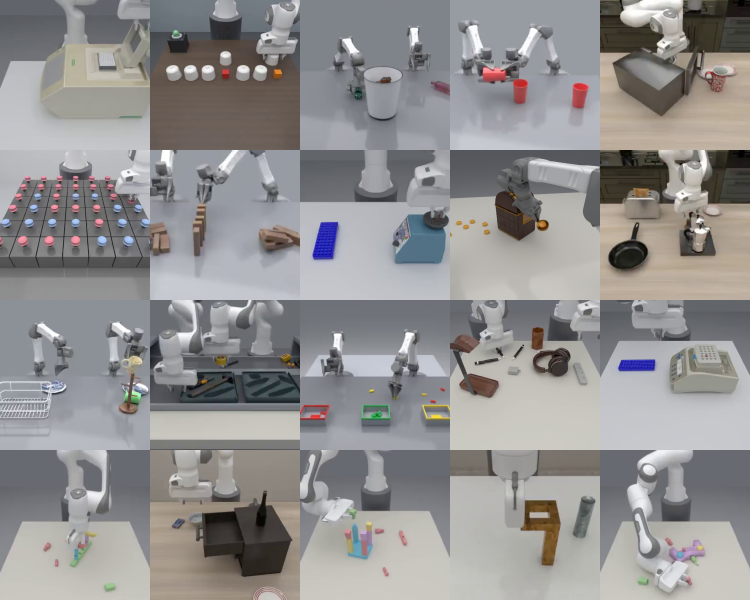}
\caption{\textbf{Representative tasks from \shortname{}.}
The benchmark includes 22 tasks spanning single and bimanual manipulation, assembly, physical puzzles, and biolab-inspired procedures. This figure shows 20 of them. Tasks require combinations of visual understanding, precise control, memory, and multi-step planning. \shortname{} supports rendering with Blender for high-fidelity visual observations.}
\label{fig:task-lineup}
\end{figure}

Recently, foundation models (FM) have begun controlling robots through several approaches: training Vision-Language-Action (VLA) models~\citep{brohan2023rt2,kim2024openvla,black2024pi0}, integrating models into hierarchical control systems~\citep{ichter2023saycan,huang2023innermonologue,belkhale2024rth,shi2025hirobot,hu2026hivla}, and generating executable perception and control code, as in Code-as-Policies (CaP)~\citep{liang2023code}, more recently inside coding-agent harnesses that write programs~\citep{fu2026capx} or computation graphs~\citep{chen2026gapgraphaspolicymultiagentselflearning}.
Frontier multimodal FMs now allow a fourth approach: the agent observes the world directly through pixels and controls robots itself with no learned policy or program in between. We refer to this setting as agentic robot control (ARC).

Evaluations of ARC disagree about how well it works. Commanding end-effector motions directly, the models in \cite{berman2026clauderobotics} completed at most 5.5\% of LIBERO tasks, or 32\% of a ten-task subset with a cursor tool. Through a 3D visual interface, VIA~\citep{hu2026via} reached 96.7\% on three LIBERO-Goal tasks. On the 42 tasks of RoboDojo, GPT-6 Astra reached 22.5\% \citep{zhang2026astrarobodojo}. Each study used its own tasks, models, and interface, so these results cannot be compared with one another. RoboARCade provides a common benchmark built for ARC that is not saturated: the VIA interface that nearly saturates LIBERO-Goal reaches 43.6\% subtask success here.

\shortname{} brings together custom and preexisting tasks and assets in a unified codebase, spanning single-arm and bimanual manipulation, assembly, physical puzzles, and high-precision biolab-inspired procedures. Fig.~\ref{fig:task-lineup} shows a subset of the task suite. For tasks adapted from prior work, we increase difficulty by broadening the range of initial conditions and improve visual fidelity through Blender rendering. All tasks run on a single simulation backend, MuJoCo, and each is a self-contained data pack, so new tasks, agents, and interfaces can be added independently.

Using this suite, we evaluate two families of frontier agents, Codex and Claude Code, and compare their performance and efficiency across two control interfaces. The first, \emph{multi-view RGBD}, provides color and depth images from multiple cameras and sends end-effector commands directly to the controller. The second, from \emph{Visual Interface Agent} (VIA)~\citep{hu2026via}, provides an interactive 3D virtual workspace in which agents can adjust a virtual target gripper using visual feedback before commanding the robot to move. The results reveal substantial differences across agents and interfaces: with VIA, GPT-5.6 Sol achieves a success rate of 14.0\%, compared with 43.6\% for Claude Fable 5.1 and 49.8\% for GPT-6 Astra. For Fable, VIA improves success over multi-view RGBD by 6.8 percentage points but increases the average cost per trial by 67\%. Even the strongest combination, Astra with VIA, succeeds on fewer than half of trials, leaving substantial room for improvement in both agents and control interfaces.
Scene understanding is rarely the obstacle: only 2 to 6\% of attempted subtasks end there. Failures come instead from choosing a poor manipulation strategy or executing a good one imprecisely. Retries help with both: Astra’s subtask success rises from 34\% on the first try to 50\% after five or more. Agents also misjudge their own progress, incorrectly claiming success on 24\% (Astra) to 39\% (Claude Opus 5) of attempted subtasks. 

We further evaluate in-context learning from successful rollouts and autonomous text-only prompt optimization on a subset of five tasks. In prompt optimization, agents revise their instructions using rollout feedback. GPT-6 Astra's mean rubric score increases from 53.5\% to 73.0\% with in-context learning and 72.0\% after two rounds of prompt optimization. Despite these gains, precision errors and incorrect claims of success persist.

\section{Related Works}

\begin{figure}[t]
\centering
\includegraphics[width=\linewidth]{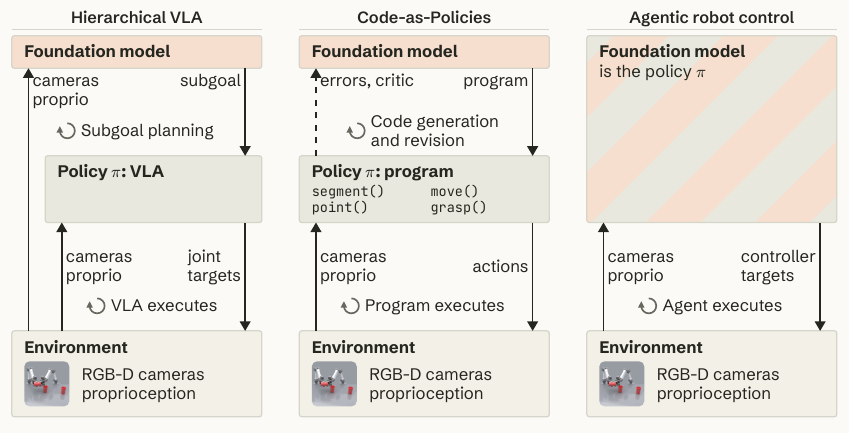}
\caption{\textbf{Different settings for robot control with FMs.}
\textbf{Hierarchical VLA:} a high-level VLM issues language subgoals to a VLA policy.
\textbf{Code-as-Policies:} an FM generates a program that controls the robot through perception and action primitives.
The FM is not involved during policy execution.
\textbf{Agentic robot control (ARC):} the FM itself is the policy, observing the scene and issuing controller targets in a closed loop.
This work focuses on the ARC setting.}
\label{fig:setups}
\end{figure}

\textbf{Foundation Models for Robot Control.}
There are many ways to leverage the general knowledge of foundation models for robot control.
Here, we briefly review the most related ones that turn an LLM or multimodal VLM into robot control policies to help position our benchmark and findings against existing ones.
\textit{Vision-Language-Action} (VLA) models are built by fine-tuning large FMs on large amounts of robot action data to control robots end-to-end~\citep{brohan2023rt2,kim2024openvla,black2024pi0}.
Early \textit{hierarchical approaches} such as SayCan~\citep{ichter2023saycan} and Inner Monologue~\citep{huang2023innermonologue} use LLMs to select and sequence low-level robot skills.
RT-H~\citep{belkhale2024rth} predicts language descriptions of low-level motions as an intermediate step between task instructions and robot actions.
More recent approaches~\citep{shi2025hirobot,hu2026hivla} add a high-level VLM for reasoning and long-term planning that issues language commands to steer a VLA, mimicking the System 1 and System 2 distinction in human cognition~\citep{evans2013dual}.
\textit{Code-as-Policies} (CaP)~\citep{liang2023code,fu2026capx,chen2026gapgraphaspolicymultiagentselflearning} uses FMs to write code given a task and a rough scene description.
In CaP-style approaches, the robot policy is the generated program, and this policy relies on perception primitives like \texttt{segment} and action primitives like \texttt{grasp} powered by neural networks or scripted functions.
In CaP-X, the coding FM itself is not involved during execution of each generated program.
It generates and revises programs between execution turns using feedback from the updated scene.

As FMs improve at multimodal understanding and agents become better at using execution feedback to guide subsequent actions, recent works explore, under different names, a setting where a model or agent acts as the policy itself to control robots directly~\citep{tsui2026demofree,berman2026clauderobotics,hu2026via,zhang2026astrarobodojo}.
We call this setting agentic robot control (ARC).
In its naive form, the agent observes the world directly through pixels and commands the robot by outputting controller targets, such as end-effector poses and the gripper open-close state for an operational-space controller.
Visual Interface Agent (VIA)~\citep{hu2026via} defines an interactable visual interface that allows agents to probe additional information from the scene by mouse-like operations and visually stage different target poses by operating a virtual target gripper inside a 3D point cloud reconstructed from RGB-D cameras, with raw camera feeds shown alongside.
The interface is a virtual workspace analogous to a GUI on a computer.
We operate in the ARC setting as it is the most straightforward way to evaluate agents' ability to control robots without relying on extra perception or action primitives.

\textbf{Simulated Robotic Benchmarks.}
Over the years, many simulated robotic benchmarks have been proposed to measure progress in robotics.
These include established suites such as Meta-World~\citep{yu2020metaworld} and RLBench~\citep{james2020rlbench}, more recent suites such as LIBERO~\citep{liu2023libero} and RoboCasa365~\citep{nasiriany2026robocasa365}, and specialized benchmarks such as RoboTwin~2.0~\citep{chen2025robotwin2} for bimanual manipulation and SoftGym~\citep{lin2021softgym} for deformable object manipulation.
ALFRED~\citep{shridhar2020alfred} evaluates household instruction following.
EmbodiedBench~\citep{yang2025embodiedbench} evaluates multimodal language model agents across household, navigation, and manipulation tasks, but its manipulation tasks supply the 3D position of every object as text and cap each episode at 15 steps, whereas \shortname{} requires the agent to localize targets from its own observations over long horizons.
CaP-X~\citep{fu2026capx} evaluates FMs using CaP and scaffolds built around vanilla CaP.
In \shortname{}, we focus on the simpler ARC setting. We evaluate agents both on their ability to complete robot control tasks and on their ability to perform in-context learning and prompt optimization to improve the task success rate.

\textbf{Benchmarks for FMs and Agents.}
Benchmarks also play an important role in the development and evaluation of FMs and agents.
Visual QA benchmarks such as BLINK~\citep{fu2024blink} evaluate FMs' visual perception in images.
Video benchmarks evaluate spatial understanding in VSI-Bench~\citep{yang2025vsibench}, temporal understanding in MVBench~\citep{li2024mvbench} and TempCompass~\citep{liu2024tempcompass}, and understanding of longer video sequences in EgoSchema~\citep{mangalam2023egoschema} and Video-MME~\citep{fu2025videomme}.
For 3D scene understanding, ScanQA~\citep{azuma2022scanqa} evaluates object-grounded question answering in indoor scans, while SQA3D~\citep{ma2023sqa3d} evaluates reasoning about a scene from an agent's specified position and orientation.
Computer-use benchmarks evaluate agents' ability to perform closed-loop control through graphical user interfaces (GUIs), including desktop applications in OSWorld~\citep{xie2024osworld}, websites in WebArena~\citep{zhou2024webarena} and VisualWebArena~\citep{koh2024visualwebarena}, and mobile applications in AndroidWorld~\citep{rawles2025androidworld}.
\shortname{} is a valuable addition to existing agentic benchmarks as its ARC tasks evaluate agents on the combinations of the aforementioned abilities including 2D and 3D visual understanding, active perception, long-term planning with visual feedback, and closed-loop control of robots.

\textbf{In-Context Learning and Prompt Optimization.}
Agents' abilities to follow instructions open up new ways for them to improve without gradient-based training.
In-context learning (ICL) and reflection-based methods allow agents to condition on prior experiences or examples included in their input to improve performance~\citep{brown2020fewshot,shinn2023reflexion,zhang2025ace}.
Prompt optimization methods improve the instructions and examples supplied to FMs through techniques such as example bootstrapping, textual feedback, and evolutionary search~\citep{khattab2024dspy,yuksekgonul2025textgrad,agrawal2026gepa}.
In robotics, prior work has explored using FMs to generate reward functions~\citep{ma2024eureka} or simulation tasks~\citep{wang2024gensim} for policy learning.
In \shortname{}, we evaluate how ICL and prompt optimization by the agent itself improve downstream robot control in the ARC setting without fine-tuning model weights.

\section{\shortname}
In this section, we first describe the benchmark infrastructure in Sec.~\ref{ssub:bi}, then its features in Sec.~\ref{ssub:features}, and finally our evaluation of autonomous improvement in Sec.~\ref{ssub:evalsi}.

\subsection{Benchmark Infrastructure}
\label{ssub:bi}

The benchmark has three components: a task suite, a simulation environment, and an agent interface. For each evaluation, we initialize the selected task and seed in the simulation environment and create an agent sandbox. The agent cannot access privileged state, such as object poses, from the sandbox. It acts in the simulator and receives observations through the selected interface.

\textbf{Simulation Environment.} Each task specifies a MuJoCo scene, instructions, robot and camera configurations, randomized initial conditions, and success predicates. For each trial, we enforce a timeout budget for each trial. The agent uses a selected interface to observe the scene and command the robot. Its commands are converted into absolute end-effector poses and then into joint-position targets for execution in MuJoCo. After each action, persistent Blender workers render high-fidelity RGB images. Interactions and videos are recorded for offline analysis, while success predicates outside the agent sandbox determine rubric scores.

\textbf{Agent Interfaces.} \textit{VIA} \citep{hu2026via} is a browser-based 3D interface with a fused point cloud, selectable camera views, and a virtual target gripper. The agent uses structured tools to select image locations, adjust the target gripper’s position and orientation, preview reachability, and execute waypoints. \textit{Multi-view RGBD} instead provides calibrated color images and aligned, quantized depth images, together with camera intrinsics, extrinsics, and end-effector proprioception. The agent specifies Cartesian translation and rotation increments and gripper commands, which execute immediately and return updated observations. Both interfaces use the same simulation and motion controller. They differ in how they present spatial information and specify actions.

\textbf{Agent Sandbox.} The agent receives task instructions, interface documentation, camera observations, and gripper readouts. It cannot access privileged simulator state, task assets, success predicates, or grading results, and it cannot modify the interface or benchmark files. Network access is limited to the services needed to use the interface and model provider. These restrictions keep task outcomes and grading information outside the agent’s view.

\textbf{Task Curation.}
We curate 22 tasks spanning precise manipulation, memory, planning, and bimanual control. For example, the Shell Game requires tracking objects through swaps, Lights Out requires planning across 42 buttons, and Strongbox and Pack Tools require completing many sequential subtasks. Furthermore, we include tasks that test precise placement under tight tolerances. Fig. \ref{fig:task-lineup} visualizes a subset of the full benchmark. The benchmark combines assets and tasks from existing open-source simulation benchmarks in addition to original assets. Exact sources for all assets in tasks are credited in the Appendix~Table \ref{tab:task-sources}.

\subsection{Features of \shortname}
\label{ssub:features}

\textbf{Simple design with photorealistic rendering.} \shortname{} is built directly on the latest version of \verb|mujoco| with no other simulation dependencies, and all task assets are hosted on Hugging Face. The agent's camera feeds are rendered in Blender at every waypoint, so the agent sees photorealistic images. Each task is able to declare its physics backend: rigid tasks run in \verb|mujoco|, and tasks with deformable objects can use Project Superdex behind the same agent interface.

\textbf{Prioritize long-horizon tasks with multiple embodiments.} Across our 22 tasks, a rollout runs for an average of 3,343 simulation steps (about 167 s at 20 Hz), while the average trajectory length is 276 steps for LIBERO-Long, 372 for RoboCasa, 502 for VLABench and 2,972 for RoboCerebra~\cite{robocerebra}, and RoboDojo's long-horizon demonstrations average 630 steps at 25 Hz~\cite{robodojo}. RoboCerebra~\cite{robocerebra} primarily tests the single Franka Panda embodiment. Our tasks test the single Franka Panda, bimanual Panda and bimanual YAM embodiments.

\textbf{Prioritize real-world tasks and easy extension.} Half of the tasks are drawn from real settings: wet-lab procedures (setting up a PCR run, loading an incubator, vortexing a sample), household routines (tidying a tabletop, making breakfast) and factory assembly (tool packing on conveyor belt). Every task is a self-contained data pack (scene, objects, language instruction, randomization and success predicates), so a new task can be easily added.

\textbf{Significant initial state randomization within tasks.} Tasks substantially randomize each object's initial placement while allowing the task to remain feasible. They also vary task-specific conditions, such as the Lights Out board color state and Shell Game random shuffling.

\textbf{Interface and model agnostic.} Interchangeable MCP interfaces let us evaluate the same tasks with either a visual 3D interface or minimal multi-view RGBD observations. A shared protocol and grader support comparisons across models and company agent harnesses, including Claude Code and Codex.

\subsection{Evaluating Autonomous Improvement}
\label{ssub:evalsi}
\textbf{In-context success demo (ICL).}
The agent first reads a successful episode of the same task, including its tool calls and images. It then attempts the task in the same session. At each action, it briefly explains how the example informed its choice.

\textbf{Multi-round task-specific text prompt optimization (T1, T2).}
The agent first distills the successful episode into a text-only \texttt{harness.md}. A fresh session uses this file as part of the prompt instead of the demonstration (text prompt, round 1, or T1). After each round, the agent revises the file using the demonstration and graded results from earlier attempts (round 2, or T2). The prompts are in Appendix~\ref{app:prompts}.

\section{Experiments}
\subsection{Experimental Setup}
We first describe the setup that both evaluations share. We evaluate \textit{agents} because Claude models run in Claude Code and GPT models run in Codex. All use the highest available reasoning effort (\texttt{xhigh}) and their runtime’s native context compaction with a custom compaction prompt (Appendix~\ref{app:prompts}). An episode ends when the agent declares the task done, or when it exhausts the task's simulation-step budget or wall-clock limit.
Each task’s rubric defines a sequence of \textit{subtasks}. After the scene comes to rest, success at each subtask is determined automatically from the simulator state using the rubric’s criteria. For example, the PCR plate setup task has four subtasks: precisely placing the plate, closing the lid, securing the latch, and turning the knob to start the machine. Throughout Sec. \ref{ssub:mrzs} and Sec. \ref{ssub:mrai}, we include direct quotes from output transcripts from different agents for how they justify their actions.

\textbf{Zero-shot Evaluation.}
We evaluate the agents GPT-5.6 Sol, Claude Opus 5, Claude Fable 5.1 and GPT-6 Astra on all 22 tasks, with the same five fixed seeds. Fable and Astra are run with two interfaces: VIA, a visual interface with a 3D point-cloud workspace and a target gripper moved by waypoint tools, and a multi-view RGBD interface that exposes multi-view RGB, depth and camera calibration and executes end-effector (EEF) commands immediately.

\textbf{Autonomous Improvement Evaluation.} We evaluate GPT-6 Astra as it performs the best in the Zero-shot Evaluation and Anthropic has strict rules that trigger \verb|reasoning_extraction|. We evaluate five tasks that have at least one full success with 10 seeds each.

\subsection{Main Results: Zero-shot Evaluation}
\label{ssub:mrzs}
\begin{table}[t]
\centering
\footnotesize
\begin{tabular}{llrrrr}
\toprule
Model & Interface & Success (\%) & Cost (\$) & Turns/EEF & Calls/EEF \\
\midrule
GPT-5.6-Sol & VIA & 14.0 & 25.05 & 3.6 & 3.3 \\
Claude Opus 5 & VIA & 30.3 & 60.00$^\dagger$ & 4.9 & 5.5 \\
\midrule
\multirow{2}{*}{Claude Fable 5.1} & Multi-view RGBD & 36.8 & 33.16 & 1.0 & 1.7 \\
 & VIA & 43.6 & 55.50 & 2.2 & 7.0 \\
\midrule
\multirow{2}{*}{GPT-6-Astra} & Multi-view RGBD & 48.3 & 13.89 & 1.3 & 1.3 \\
 & VIA & 49.8 & 31.84 & 2.3 & 3.0 \\
\bottomrule
\end{tabular}
\caption{\textbf{Benchmark on \shortname}. All metrics are averaged over 22 tasks with 5 seeds each, over all trials. Success measures completed rubric subtasks. Turns/EEF and Calls/EEF divide a trial's model turns (requests to the model, each of which can issue several tool calls) and tool calls by the number of distinct end-effector (EEF) commands the agent gives the simulator. GPT-6 Astra has the highest success rate even using a minimal multi-view RGBD interface. $^\dagger$A lower bound: two Opus trials each lost the usage record of one interrupted response. Success-only metrics, EEF command counts and wall time per command are in Appendix Table~\ref{tab:main-results-full}.}
\label{tab:main-results}
\end{table}

To explain failures beyond automatic rubric-based success detection, we classify each attempted subtask by the first step at which the agent fails: \textit{visual understanding} (what to do in the scene), \textit{approach} (how to accomplish the subtask), or \textit{precision} (precisely executing the subtask). We first use position and rotation errors to distinguish approach from precision failures. For example, a beam-placement attempt within 25~mm and $6^\circ$ of the target but outside the 0.6~mm lateral success tolerance is provisionally labeled a precision failure. A VLM or human annotator then verifies the approach–precision boundary and distinguishes visual understanding failures from approach failures.

\begin{figure}[t]
\centering
\includegraphics[width=\linewidth]{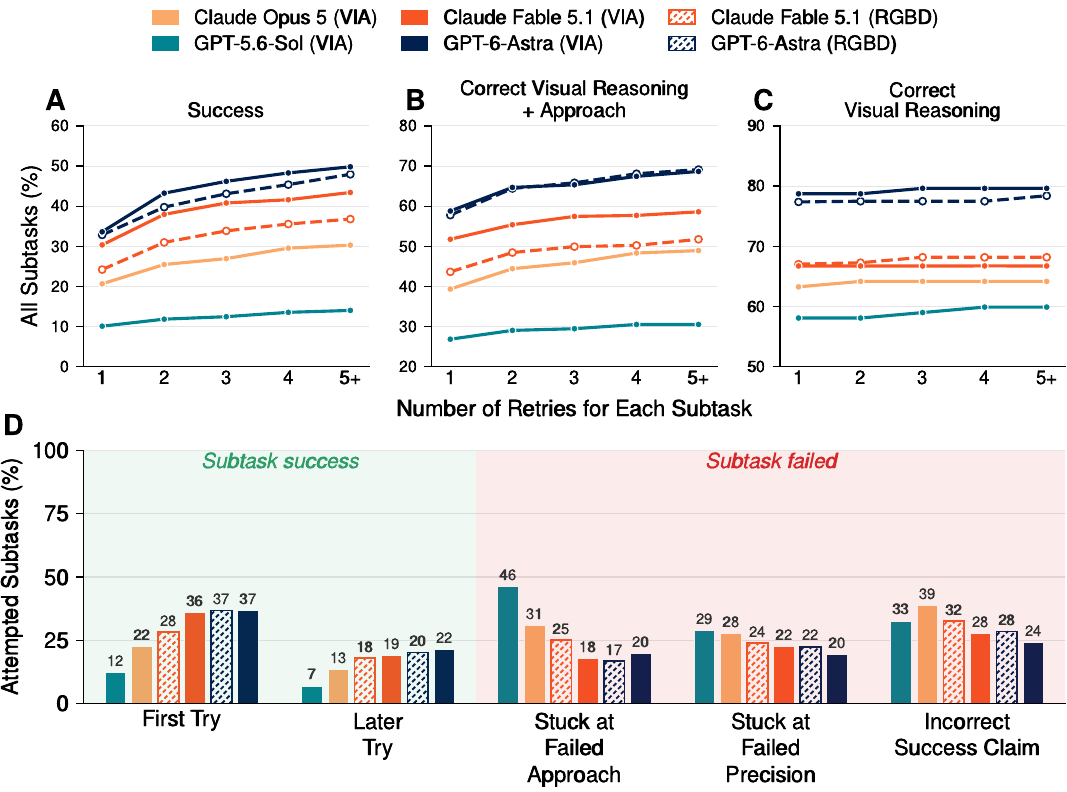}
\caption{\textbf{Failure Breakdown for \shortname. A-C}: These plots are over all rubric subtasks, including those the agent never reached, the share succeeded (A), with correct visual reasoning and approach (B), and with correct visual reasoning (C) within the first \(N\) tries at the subtask. Success requires correct visual reasoning, approach and precision. GPT-6 Astra generally performs best. More tries mainly improve approach and success. Visual reasoning changes little across agents. \textbf{D:} This plot breaks down agent behavior on subtasks it attempted. Stronger agents are more likely to succeed on later tries. When they get stuck, weaker agents fail more on approach, while stronger agents fail about as often on approach as on precision. Stronger models also make fewer incorrect success claims, though GPT-6 Astra still incorrectly claims success on 24\% of subtasks.}
\label{fig:retry-verification}
\end{figure}

\textbf{Q1: Where in a task do agents fail: visual understanding, approach, or precision?} First, as expected, stronger agents attempt more subtasks. Agents attempt only 62 to 82\% of all subtasks because they run out of budget or stall on an earlier subtask (the red line in Appendix Fig.~\ref{fig:retry-stacked}A). Of the subtasks the agents attempt, visual reasoning is rarely the final failure as only 2 to 6\% end with it (Appendix Fig.~\ref{fig:retry-stacked}B). The remaining failures split between approach and precision. Weaker agents get stuck more often on approach (Sol VIA 46\% against 29\% on precision), while the stronger agents get stuck on both about equally (Astra VIA 20\% each), and only Fable VIA and Astra RGBD get stuck more on precision, by about 5 percentage points (pp) (Fig.~\ref{fig:retry-verification}D). Agents often measure these misses themselves, as in ``the gripper is sitting 3-4mm off-center toward the -Y end" (Fable VIA) and ``the tip appears roughly 1.5 cm too high" (Astra VIA). Approach failures are incorrect grasp or non-prehensile motion plans, which the agent sometimes identifies such as ``a vertical approach cannot reach the edge cup from this arm configuration" (Astra VIA) or ``this block can't be taken from directly above" (Opus VIA). 
The interface changes how agents work toward precision: with VIA, agents make 3.0 to 7.0 tool calls and 2.2 to 4.9 model turns per end-effector command, against 1.3 to 1.7 tool calls and 1.0 to 1.3 model turns with raw RGBD (Table~\ref{tab:main-results}). They spend the extra tool calls and turns inspecting the scene before moving, which also increases the cost.

\textbf{Q2: Do agents recover from failed attempts?}
Agents mostly improve on the first retry and mostly from approach and precision errors (Fig.~\ref{fig:retry-verification}A to C). For Astra VIA, subtask success rises from 34\% within one try to 43\% within two and 50\% after five or more (Fig.~\ref{fig:retry-verification}A). Recovery depends on if the failure was visual understanding, approach, or precision (Appendix Table~\ref{tab:retry-transitions}). For all agents, a visual reasoning failure barely improves across retries: from the first to the fifth try, the share of subtasks with correct visual reasoning rises by at most 1.8~pp (Sol VIA, Fig.~\ref{fig:retry-verification}C). After a precision miss, the next try succeeds 33\% of the time for Astra and 37\% for Fable with VIA, but only 26\% and 27\% with multi-view RGBD (Appendix Table~\ref{tab:retry-transitions}). When agents make successful corrections, they are measured and small (``the opening appears a few millimeters off the peg along the robot's x axis,” Astra VIA). After a wrong approach, Fable VIA recovers most often (26\%, Appendix Table~\ref{tab:retry-transitions}), and Astra VIA succeeds at the most attempted subtasks on a later try (22\% in Fig.~\ref{fig:retry-verification}D). Sol repeats a wrong approach on 76\% of its retries after one and recovers from it only 2\% of the time (Appendix Table~\ref{tab:retry-transitions}).

\textbf{Q3: Do agents know when a subtask is done?}
Agents do not reliably know when a subtask is done. Agents incorrectly claim success on 24\% (Astra VIA) to 39\% (Opus VIA) of attempted subtasks, either ending the episode or moving to the next subtask before the success criteria are met (Incorrect Success Claim, Fig.~\ref{fig:retry-verification}D). Even in trials where the part was never placed, agents confidently declare success: ``the pink crossbar is seated over the blue peg...and stable after release” (Astra VIA) and ``the nut is lying flat around the peg base...confirming the task is complete” (Fable VIA). VIA reduces incorrect claims relative to multiview RGBD by 5~pp for Fable and 4~pp for Astra (Fig.~\ref{fig:retry-verification}D).

\textbf{Q4: What differs across models and interfaces?}
Astra is both the strongest and the cheaper model with either interface: with minimal RGBD it reaches 48.3\% success against 36.8\% for Fable, and with VIA 49.8\% against 43.6\%, at a lower cost per trial in both cases (Table~\ref{tab:main-results}). Among the VIA models, Fable and Astra use similar model turns per EEF command but Astra performs better (Table~\ref{tab:main-results}). Worse agents like Opus and Sol have the lower success rates but the highest model turns (Table~\ref{tab:main-results}). Turning to the interface, VIA increases cost compared to the minimal interface while yielding only modest gains in task success. The minimal multi-view RGBD interface uses roughly half as many model turns per end-effector command as VIA and costs less for both agents. VIA increases success by 6.8 percentage points for Fable and 1.5 points for Astra (Table~\ref{tab:main-results}). As discussed in Q3, VIA mainly helps agents reduce incorrect success claims.

\subsection{Main Results: Autonomous Improvement Evaluation}
\label{ssub:mrai}
\textbf{Q5: Does context increase success rates and why?}
When given a successful episode, the agent uses it as a task-specific procedure. In the baseline, many failures exhaust the budget after the agent stalls on an incorrect approach. Adding one successful Astra episode of the same task to context (ICL) reduces these timeouts and raises the mean rubric score from 53.5\% to 73.0\% (Table~\ref{tab:auto-improve}). Before acting, the agent draws on the demonstration’s procedure, noting in the bimanual tasks that ``an isolated inner-left block was a reliable first pickup, and six-centimeter spacing supported the full chain" and that ``the live plate is angled... so I'll twist the jaws to match it." Since the initial demo loading is expensive, we can test how much of a successful demonstration survives distillation into text. A text prompt written from the same episode (T1) reaches a 62.5\% mean rubric score. Revising it with the graded T1 trials (T2) raises the score to 72.0\%, nearly matching the 73.0\% of ICL, but requires additional rollouts (Table~\ref{tab:auto-improve}). Baseline and ICL trial costs are similar: the baseline uses more end-effector commands (Appendix Table~\ref{tab:auto-improve-full}), while the demonstration adds upfront cost. T2 is cheapest per trial because it uses fewer model turns (Appendix Table~\ref{tab:auto-improve-full}), while avoiding the cost of loading the full successful episode into context. Another benefit of a text-only prompt is that it can be written from prior attempts even when no successful demonstration is available. On a task Astra failed in all five baseline trials, a prompt written from those failures achieves 60\% success (Table~\ref{tab:auto-improve}). Revising it with the five T1 trials raises success to 5 of 5. Cost per trial falls from \$48.00 at baseline to \$30.27 with T1 and \$27.20 with T2 (Table~\ref{tab:auto-improve}).

\textbf{Q6: What kind of failures persist?} 
The three main failures are precision errors, incorrect success claims, and situations outside the demonstration. (i) Precision errors: the two tasks with precise placement have similar success rates under every method (Appendix Table~\ref{tab:auto-improve-tasks}). (ii) Incorrect success claims: Most remaining failures with ICL and T2 end when the agent incorrectly declares success. The number is similar across methods, with 12 such trials under ICL, 13 under T2, and 12 at baseline (Appendix Table~\ref{tab:auto-improve-tasks}). (iii) Situations outside the demonstration: when the scene deviates from the demonstration, the agent must improvise. In one rollout, it reports no similarity to the demonstration: "None. The blue post remains securely held, but the yellow beam has come loose."

\begin{table}[t]
\centering
\footnotesize
\begin{tabular}{lrrrr}
\toprule
Method & Success (\%) & Cost (\$) & Turns/EEF & Calls/EEF \\
\midrule
GPT-6-Astra baseline & 53.5 & 27.60 & 2.5 & 3.4 \\
In-context success demo (ICL) & 73.0 & 30.55 & 3.1 & 4.2 \\
Text prompt, round 1 (T1) & 62.5 & 30.76 & 3.0 & 3.3 \\
Text prompt, round 2 (T2) & 72.0 & 22.75 & 2.3 & 3.2 \\
\specialrule{1.2pt}{2pt}{2pt}
\multicolumn{5}{l}{\textit{No successful episode available}} \\
GPT-6-Astra baseline & 0.0 & 48.00 & 3.6 & 4.1 \\
Text prompt, round 1 (T1) & 60.0 & 30.27 & 1.8 & 4.8 \\
Text prompt, round 2 (T2) & 100.0 & 27.20 & 3.5 & 4.3 \\
\bottomrule
\end{tabular}
\caption{\textbf{Autonomous improvement on \shortname}. \textbf{Top:} We evaluate GPT-6-Astra where metrics are averaged over five partially successful tasks with 10 trials per task and method. ICL loads a full successful episode before the agent attempts the task and raises the mean rubric score from 53.5\% to 73.0\%. A text-only prompt reaches 72.0\% after two rounds of revision using graded rollouts (T2). \textbf{Bottom:} For a task with no successful Astra episode, T1 prompt is written from the five failed baseline trials and the T2 prompt from those and the five T1 trials, evaluated on 5 fresh trials each. Although, a small sample size, T2 is able to achieve 100\% performance for a task that had zero full successes. Full metrics are in Appendix Table~\ref{tab:auto-improve-full}.}
\label{tab:auto-improve}
\end{table}





\section{Conclusion}
We introduced \shortname{}, a benchmark of 22 long-horizon tabletop manipulation tasks with wide initial-state randomization and photorealistic rendering. We evaluate agents through two interfaces in a sandbox that hides privileged state, both zero-shot and after autonomous improvement. Zero-shot, GPT-6 Astra is the strongest agent with either interface and costs less per trial than Claude Fable 5.1, reaching 49.8\% subtask success with VIA and 48.3\% with minimal multi-view RGBD. Thus, as agents continue to improve additional harnesses might be less relevant. Across agents, failures stem more often from approach, precision, and failure to recognize completion than from visual misunderstanding. Retries recover approach and precision but even stronger models still incorrectly claims success on attempted subtasks (Astra at 24\%). Agents can also improve themselves without human input: one successful episode in context raises Astra's mean rubric score from 53.5\% to 73.0\%, a text prompt the agent distills from that episode and revises with graded rollouts reaches 72.0\% at lower cost, and a prompt written from failed attempts alone can improve a task's success rate. What context does not fix is precision, incorrect success claims, and situations the demonstration did not cover. This benchmark analyzes the failure cases of state-of-the-art agents on simulated tasks, and it takes a first step toward benchmarking autonomous improvement for agentic robotic control.

\textbf{Limitations:} Constrained to practical cost considerations, we selected a set of tasks and five seeds to evaluate. For both the zero-shot and autonomous improvement evaluations, we plan to run more seeds and expand the task set to include mobile manipulation and dexterous manipulation with robot hands. Also, the existing interface only allow for single EEF commands and these tasks are mostly quasi-static but it would be exciting to explore interfaces with full EEF trajectories to accomplish more dynamic tasks.
\section{Acknowledgements}
We thank IPRL and ILIAD members for insightful discussions. In particular, we thank Samrat Sahoo
for help with the task development. We also thank Sarthak Kamat for helpful general discussions about the project. 
Satvik Sharma is supported by the Stanford Robotics Center Robotics Scholars Program. This project
was supported by Meta, PECASE, ONR MURI N00014-25-1-2479  and the National Science Foundation (NSF) under Grant Numbers 2342246, 2327974, 2132847, 1941722.
\clearpage
\section*{AI Use Statement}

We primarily used generative AI tools for writing and editing, literature assistance, creating and modifying figures, and coding assistance. LLM-generated code sections were verified and tested for correctness by one to two authors. Figures made with AI tools were iterated on by multiple authors. The authors used AI for editing sections which were first formulated and outlined then edited with AI. The authors themselves designed the method and experiments. We take responsibility for the final content of this work, including all AI-assisted text, code, and figures.

\section*{Reproducibility Statement}

The paper specifies the necessary details to rerun our experiments, and we have used only publicly accessible models and environments in our work.
Exact reproduction of the rollouts is impossible due to the non-deterministic nature of large foundation model inference via API calls, but the general trends and aggregated behavior across seeds can be reproduced.
We will release the benchmark code and evaluation scripts to support reproduction of our results.

\bibliography{iclr2026_conference}

@inproceedings{liang2023code,
  title     = {{Code as Policies: Language Model Programs for Embodied Control}},
  author    = {Liang, Jacky and Huang, Wenlong and Xia, Fei and Xu, Peng and Hausman, Karol and Ichter, Brian and Florence, Pete and Zeng, Andy},
  booktitle = {IEEE International Conference on Robotics and Automation (ICRA)},
  year      = {2023},
}

@article{hu2026via,
  title   = {{{VIA}: Visual Interface Agent for Robot Control}},
  author  = {Hu, Hengyuan and Sundaresan, Priya and Gao, Jensen and Sadigh, Dorsa},
  journal = {arXiv preprint arXiv:2607.11119},
  year    = {2026},
}

@inproceedings{fu2026capx,
  title     = {{{CaP-X}: A Framework for Benchmarking and Improving Coding Agents for Robot Manipulation}},
  author    = {Fu, Letian and Yu, Justin and El-Refai, Karim and Kou, Ethan and Xue, Haoru and Huang, Huang and Xiao, Wenli and Wang, Guanzhi and Niu, Dantong and Li, Fei-Fei and Shi, Guanya and Wu, Jiajun and Sastry, Shankar and Zhu, Yuke and Goldberg, Ken and Fan, Linxi},
  booktitle = {International Conference on Machine Learning (ICML)},
  year      = {2026},
}

@article{belkhale2024rth,
  title   = {{{RT-H}: Action Hierarchies Using Language}},
  author  = {Belkhale, Suneel and Ding, Tianli and Xiao, Ted and Sermanet, Pierre and Vuong, Quan and Tompson, Jonathan and Chebotar, Yevgen and Dwibedi, Debidatta and Sadigh, Dorsa},
  journal = {arXiv preprint arXiv:2403.01823},
  year    = {2024},
  url     = {https://arxiv.org/abs/2403.01823},
}

@inproceedings{ichter2023saycan,
  title     = {{Do As I Can, Not As I Say: Grounding Language in Robotic Affordances}},
  author    = {Ichter, Brian and Brohan, Anthony and Chebotar, Yevgen and Finn, Chelsea and Hausman, Karol and Herzog, Alexander and Ho, Daniel and Ibarz, Julian and Irpan, Alex and Jang, Eric and Julian, Ryan and Kalashnikov, Dmitry and Levine, Sergey and Lu, Yao and Parada, Carolina and Rao, Kanishka and Sermanet, Pierre and Toshev, Alexander T. and Vanhoucke, Vincent and Xia, Fei and Xiao, Ted and Xu, Peng and Yan, Mengyuan and Brown, Noah and Ahn, Michael and Cortes, Omar and Sievers, Nicolas and Tan, Clayton and Xu, Sichun and Reyes, Diego and Rettinghouse, Jarek and Quiambao, Jornell and Pastor, Peter and Luu, Linda and Lee, Kuang-Huei and Kuang, Yuheng and Jesmonth, Sally and Joshi, Nikhil J. and Jeffrey, Kyle and Ruano, Rosario Jauregui and Hsu, Jasmine and Gopalakrishnan, Keerthana and David, Byron and Zeng, Andy and Fu, Chuyuan Kelly},
  booktitle = {Conference on Robot Learning (CoRL)},
  pages     = {287--318},
  year      = {2023},
  url       = {https://proceedings.mlr.press/v205/ichter23a.html},
}

@inproceedings{huang2023innermonologue,
  title     = {{Inner Monologue: Embodied Reasoning through Planning with Language Models}},
  author    = {Huang, Wenlong and Xia, Fei and Xiao, Ted and Chan, Harris and Liang, Jacky and Florence, Pete and Zeng, Andy and Tompson, Jonathan and Mordatch, Igor and Chebotar, Yevgen and Sermanet, Pierre and Jackson, Tomas and Brown, Noah and Luu, Linda and Levine, Sergey and Hausman, Karol and Ichter, Brian},
  booktitle = {Conference on Robot Learning (CoRL)},
  pages     = {1769--1782},
  year      = {2023},
  url       = {https://proceedings.mlr.press/v205/huang23c.html},
}

@inproceedings{shi2025hirobot,
  title     = {{{Hi Robot}: Open-Ended Instruction Following with Hierarchical Vision-Language-Action Models}},
  author    = {Shi, Lucy Xiaoyang and Ichter, Brian and Equi, Michael Robert and Ke, Liyiming and Pertsch, Karl and Vuong, Quan and Tanner, James and Walling, Anna and Wang, Haohuan and Fusai, Niccolo and Li-Bell, Adrian and Driess, Danny and Groom, Lachy and Levine, Sergey and Finn, Chelsea},
  booktitle = {International Conference on Machine Learning (ICML)},
  pages     = {54919--54933},
  year      = {2025},
  url       = {https://proceedings.mlr.press/v267/shi25d.html},
}

@article{hu2026hivla,
  title   = {{What Matters in Orchestrating Robot Policies: A Systematic Study of Hierarchical {VLA} Agents}},
  author  = {Hu, Jiaheng and Shridhar, Mohit and Lu, Caden and Shah, Dhruv and Chiang, Hao-Tien Lewis and Tan, Jie and Xie, Annie},
  journal = {arXiv preprint arXiv:2606.10267},
  year    = {2026},
  url     = {https://arxiv.org/abs/2606.10267},
}

@inproceedings{brohan2023rt2,
  title     = {{{RT-2}: Vision-Language-Action Models Transfer Web Knowledge to Robotic Control}},
  author    = {Zitkovich, Brianna and others},
  booktitle = {Conference on Robot Learning (CoRL)},
  year      = {2023},
}

@inproceedings{kim2024openvla,
  title     = {{{OpenVLA}: An Open-Source Vision-Language-Action Model}},
  author    = {Kim, Moo Jin and Pertsch, Karl and Karamcheti, Siddharth and Xiao, Ted and Balakrishna, Ashwin and Nair, Suraj and Rafailov, Rafael and Foster, Ethan P. and Sanketi, Pannag R. and Vuong, Quan and Kollar, Thomas and Burchfiel, Benjamin and Tedrake, Russ and Sadigh, Dorsa and Levine, Sergey and Liang, Percy and Finn, Chelsea},
  booktitle = {Conference on Robot Learning (CoRL)},
  year      = {2025},
}

@inproceedings{black2024pi0,
  title     = {{$\pi_0$: A {Vision-Language-Action} Flow Model for General Robot Control}},
  author    = {Black, Kevin and Brown, Noah and Driess, Danny and Esmail, Adnan and Equi, Michael Robert and Finn, Chelsea and Fusai, Niccolo and Groom, Lachy and Hausman, Karol and Ichter, Brian and Jakubczak, Szymon and Jones, Tim and Ke, Liyiming and Levine, Sergey and Li-Bell, Adrian and Mothukuri, Mohith and Nair, Suraj and Pertsch, Karl and Shi, Lucy Xiaoyang and Smith, Laura and Tanner, James and Vuong, Quan and Walling, Anna and Wang, Haohuan and Zhilinsky, Ury},
  booktitle = {Robotics: Science and Systems (RSS)},
  year      = {2025},
}

@inproceedings{yu2020metaworld,
  title     = {{{Meta-World}: A Benchmark and Evaluation for Multi-Task and Meta Reinforcement Learning}},
  author    = {Yu, Tianhe and Quillen, Deirdre and He, Zhanpeng and Julian, Ryan and Hausman, Karol and Finn, Chelsea and Levine, Sergey},
  booktitle = {Conference on Robot Learning (CoRL)},
  pages     = {1094--1100},
  year      = {2020},
  url       = {https://proceedings.mlr.press/v100/yu20a.html},
}

@inproceedings{nasiriany2026robocasa365,
  title     = {{{RoboCasa365}: A Large-Scale Simulation Framework for Training and Benchmarking Generalist Robots}},
  author    = {Nasiriany, Soroush and Nasiriany, Sepehr and Maddukuri, Abhiram and Zhu, Yuke},
  booktitle = {International Conference on Learning Representations (ICLR)},
  year      = {2026},
  url       = {https://robocasa.ai/},
}

@article{chen2025robotwin2,
  title   = {{{RoboTwin 2.0}: A Scalable Data Generator and Benchmark with Strong Domain Randomization for Robust Bimanual Robotic Manipulation}},
  author  = {Chen, Tianxing and Chen, Zanxin and Chen, Baijun and Cai, Zijian and Liu, Yibin and Li, Zixuan and Liang, Qiwei and Lin, Xianliang and Ge, Yiheng and Gu, Zhenyu and Deng, Weiliang and Guo, Yubin and Nian, Tian and Xie, Xuanbing and Chen, Qiangyu and Su, Kailun and Xu, Tianling and Liu, Guodong and Hu, Mengkang and Gao, Huan-ang and Wang, Kaixuan and Liang, Zhixuan and Qin, Yusen and Yang, Xiaokang and Luo, Ping and Mu, Yao},
  journal = {arXiv preprint arXiv:2506.18088},
  year    = {2025},
  url     = {https://arxiv.org/abs/2506.18088},
}

@inproceedings{lin2021softgym,
  title     = {{{SoftGym}: Benchmarking Deep Reinforcement Learning for Deformable Object Manipulation}},
  author    = {Lin, Xingyu and Wang, Yufei and Olkin, Jake and Held, David},
  booktitle = {Conference on Robot Learning (CoRL)},
  pages     = {432--448},
  year      = {2021},
  url       = {https://proceedings.mlr.press/v155/lin21a.html},
}

@article{james2020rlbench,
  title   = {{{RLBench}: The Robot Learning Benchmark \& Learning Environment}},
  author  = {James, Stephen and Ma, Zicong and Arrojo, David Rovick and Davison, Andrew J.},
  journal = {IEEE Robotics and Automation Letters},
  volume  = {5},
  number  = {2},
  pages   = {3019--3026},
  year    = {2020},
}

@inproceedings{liu2023libero,
  title     = {{{LIBERO}: Benchmarking Knowledge Transfer for Lifelong Robot Learning}},
  author    = {Liu, Bo and Zhu, Yifeng and Gao, Chongkai and Feng, Yihao and Liu, Qiang and Zhu, Yuke and Stone, Peter},
  booktitle = {Advances in Neural Information Processing Systems (NeurIPS), Datasets and Benchmarks Track},
  year      = {2023},
}

@inproceedings{shridhar2020alfred,
  title     = {{{ALFRED}: A Benchmark for Interpreting Grounded Instructions for Everyday Tasks}},
  author    = {Shridhar, Mohit and Thomason, Jesse and Gordon, Daniel and Bisk, Yonatan and Han, Winson and Mottaghi, Roozbeh and Zettlemoyer, Luke and Fox, Dieter},
  booktitle = {IEEE/CVF Conference on Computer Vision and Pattern Recognition (CVPR)},
  year      = {2020},
}

@inproceedings{yang2025embodiedbench,
  title     = {{{EmbodiedBench}: Comprehensive Benchmarking Multi-modal Large Language Models for Vision-Driven Embodied Agents}},
  author    = {Yang, Rui and Chen, Hanyang and Zhang, Junyu and Zhao, Mark and Qian, Cheng and Wang, Kangrui and Wang, Qineng and Koripella, Teja Venkat and Movahedi, Marziyeh and Li, Manling and Ji, Heng and Zhang, Huan and Zhang, Tong},
  booktitle = {International Conference on Machine Learning (ICML)},
  year      = {2025},
}

@misc{berman2026clauderobotics,
  title        = {{Claude} plays robotics},
  author       = {Berman, Shmuel and Ilie, Michael and Deng, Jia and Freeman, Daniel},
  year         = {2026},
  month        = jul,
  howpublished = {Anthropic},
  url          = {https://www.anthropic.com/research/claude-plays-robotics},
}

@inproceedings{xie2024osworld,
  title     = {{{OSWorld}: Benchmarking Multimodal Agents for Open-Ended Tasks in Real Computer Environments}},
  author    = {Xie, Tianbao and Zhang, Danyang and Chen, Jixuan and Li, Xiaochuan and Zhao, Siheng and Cao, Ruisheng and Hua, Toh Jing and Cheng, Zhoujun and Shin, Dongchan and Lei, Fangyu and Liu, Yitao and Xu, Yiheng and Zhou, Shuyan and Savarese, Silvio and Xiong, Caiming and Zhong, Victor and Yu, Tao},
  booktitle = {Advances in Neural Information Processing Systems (NeurIPS), Datasets and Benchmarks Track},
  year      = {2024},
}

@inproceedings{zhou2024webarena,
  title     = {{{WebArena}: A Realistic Web Environment for Building Autonomous Agents}},
  author    = {Zhou, Shuyan and Xu, Frank F. and Zhu, Hao and Zhou, Xuhui and Lo, Robert and Sridhar, Abishek and Cheng, Xianyi and Ou, Tianyue and Bisk, Yonatan and Fried, Daniel and Alon, Uri and Neubig, Graham},
  booktitle = {International Conference on Learning Representations (ICLR)},
  year      = {2024},
}

@inproceedings{koh2024visualwebarena,
  title     = {{{VisualWebArena}: Evaluating Multimodal Agents on Realistic Visual Web Tasks}},
  author    = {Koh, Jing Yu and Lo, Robert and Jang, Lawrence and Duvvur, Vikram and Lim, Ming Chong and Huang, Po-Yu and Neubig, Graham and Zhou, Shuyan and Salakhutdinov, Ruslan and Fried, Daniel},
  booktitle = {Annual Meeting of the Association for Computational Linguistics (ACL)},
  year      = {2024},
  url       = {https://arxiv.org/abs/2401.13649},
}

@inproceedings{rawles2025androidworld,
  title     = {{{AndroidWorld}: A Dynamic Benchmarking Environment for Autonomous Agents}},
  author    = {Rawles, Chris and Clinckemaillie, Sarah and Chang, Yifan and Waltz, Jonathan and Lau, Gabrielle and Fair, Marybeth and Li, Alice and Bishop, William and Li, Wei and Campbell-Ajala, Folawiyo and Toyama, Daniel and Berry, Robert and Tyamagundlu, Divya and Lillicrap, Timothy and Riva, Oriana},
  booktitle = {International Conference on Learning Representations (ICLR)},
  year      = {2025},
}

@inproceedings{azuma2022scanqa,
  title     = {{{ScanQA}: {3D} Question Answering for Spatial Scene Understanding}},
  author    = {Azuma, Daichi and Miyanishi, Taiki and Kurita, Shuhei and Kawanabe, Motoaki},
  booktitle = {IEEE/CVF Conference on Computer Vision and Pattern Recognition (CVPR)},
  pages     = {19129--19139},
  year      = {2022},
  url       = {https://arxiv.org/abs/2112.10482},
}

@inproceedings{ma2023sqa3d,
  title     = {{{SQA3D}: Situated Question Answering in {3D} Scenes}},
  author    = {Ma, Xiaojian and Yong, Silong and Zheng, Zilong and Li, Qing and Liang, Yitao and Zhu, Song-Chun and Huang, Siyuan},
  booktitle = {International Conference on Learning Representations (ICLR)},
  year      = {2023},
  url       = {https://openreview.net/forum?id=IDJx97BC38},
}

@inproceedings{fu2024blink,
  title     = {{{BLINK}: Multimodal Large Language Models Can See but Not Perceive}},
  author    = {Fu, Xingyu and Hu, Yushi and Li, Bangzheng and Feng, Yu and Wang, Haoyu and Lin, Xudong and Roth, Dan and Smith, Noah A. and Ma, Wei-Chiu and Krishna, Ranjay},
  booktitle = {European Conference on Computer Vision (ECCV)},
  year      = {2024},
}

@inproceedings{yang2025vsibench,
  title     = {{Thinking in Space: How Multimodal Large Language Models See, Remember, and Recall Spaces}},
  author    = {Yang, Jihan and Yang, Shusheng and Gupta, Anjali W. and Han, Rilyn and Fei-Fei, Li and Xie, Saining},
  booktitle = {IEEE/CVF Conference on Computer Vision and Pattern Recognition (CVPR)},
  year      = {2025},
}

@inproceedings{mangalam2023egoschema,
  title     = {{{EgoSchema}: A Diagnostic Benchmark for Very Long-form Video Language Understanding}},
  author    = {Mangalam, Karttikeya and Akshulakov, Raiymbek and Malik, Jitendra},
  booktitle = {Advances in Neural Information Processing Systems (NeurIPS), Datasets and Benchmarks Track},
  year      = {2023},
}

@inproceedings{li2024mvbench,
  title     = {{{MVBench}: A Comprehensive Multi-modal Video Understanding Benchmark}},
  author    = {Li, Kunchang and Wang, Yali and He, Yinan and Li, Yizhuo and Wang, Yi and Liu, Yi and Wang, Zun and Xu, Jilan and Chen, Guo and Luo, Ping and Wang, Limin and Qiao, Yu},
  booktitle = {IEEE/CVF Conference on Computer Vision and Pattern Recognition (CVPR)},
  year      = {2024},
}

@inproceedings{liu2024tempcompass,
  title     = {{{TempCompass}: Do Video {LLMs} Really Understand Videos?}},
  author    = {Liu, Yuanxin and Li, Shicheng and Liu, Yi and Wang, Yuxiang and Ren, Shuhuai and Li, Lei and Chen, Sishuo and Sun, Xu and Hou, Lu},
  booktitle = {Findings of the Association for Computational Linguistics (ACL)},
  year      = {2024},
}

@inproceedings{fu2025videomme,
  title     = {{{Video-MME}: The First-Ever Comprehensive Evaluation Benchmark of Multi-modal {LLMs} in Video Analysis}},
  author    = {Fu, Chaoyou and Dai, Yuhan and Luo, Yongdong and Li, Lei and Ren, Shuhuai and Zhang, Renrui and Wang, Zihan and Zhou, Chenyu and Shen, Yunhang and Zhang, Mengdan and Chen, Peixian and Li, Yanwei and Lin, Shaohui and Zhao, Sirui and Li, Ke and Xu, Tong and Zheng, Xiawu and Chen, Enhong and Shan, Caifeng and He, Ran and Sun, Xing},
  booktitle = {IEEE/CVF Conference on Computer Vision and Pattern Recognition (CVPR)},
  year      = {2025},
}

@inproceedings{brown2020fewshot,
  title     = {{Language Models are Few-Shot Learners}},
  author    = {Brown, Tom B. and others},
  booktitle = {Advances in Neural Information Processing Systems (NeurIPS)},
  year      = {2020},
  url       = {https://proceedings.neurips.cc/paper/2020/hash/1457c0d6bfcb4967418bfb8ac142f64a-Abstract.html},
}

@inproceedings{khattab2024dspy,
  title     = {{{DSPy}: Compiling Declarative Language Model Calls into State-of-the-Art Pipelines}},
  author    = {Khattab, Omar and Singhvi, Arnav and Maheshwari, Paridhi and Zhang, Zhiyuan and Santhanam, Keshav and Vardhamanan, Sri and Haq, Saiful and Sharma, Ashutosh and Joshi, Thomas T. and Moazam, Hanna and Miller, Heather and Zaharia, Matei and Potts, Christopher},
  booktitle = {International Conference on Learning Representations (ICLR)},
  year      = {2024},
}

@article{yuksekgonul2025textgrad,
  title   = {{Optimizing generative {AI} by backpropagating language model feedback}},
  author  = {Yuksekgonul, Mert and Bianchi, Federico and Boen, Joseph and Liu, Sheng and Lu, Pan and Huang, Zhi and Guestrin, Carlos and Zou, James},
  journal = {Nature},
  volume  = {639},
  pages   = {609--616},
  year    = {2025},
}

@inproceedings{agrawal2026gepa,
  title     = {{{GEPA}: Reflective Prompt Evolution Can Outperform Reinforcement Learning}},
  author    = {Agrawal, Lakshya A. and Tan, Shangyin and Soylu, Dilara and Ziems, Noah and Khare, Rishi and Opsahl-Ong, Krista and Singhvi, Arnav and Shandilya, Herumb and Ryan, Michael J. and Jiang, Meng and Potts, Christopher and Sen, Koushik and Dimakis, Alexandros G. and Stoica, Ion and Klein, Dan and Zaharia, Matei and Khattab, Omar},
  booktitle = {International Conference on Learning Representations (ICLR)},
  year      = {2026},
  url       = {https://arxiv.org/abs/2507.19457},
}

@article{zhang2025ace,
  title   = {{Agentic Context Engineering: Evolving Contexts for Self-Improving Language Models}},
  author  = {Zhang, Qizheng and Hu, Changran and Upasani, Shubhangi and Ma, Boyuan and Hong, Fenglu and Kamanuru, Vamsidhar and Rainton, Jay and Wu, Chen and Ji, Mengmeng and Li, Hanchen and Thakker, Urmish and Zou, James and Olukotun, Kunle},
  journal = {arXiv preprint arXiv:2510.04618},
  year    = {2025},
  url     = {https://arxiv.org/abs/2510.04618},
}

@inproceedings{shinn2023reflexion,
  title     = {{Reflexion: Language Agents with Verbal Reinforcement Learning}},
  author    = {Shinn, Noah and Cassano, Federico and Gopinath, Ashwin and Narasimhan, Karthik and Yao, Shunyu},
  booktitle = {Advances in Neural Information Processing Systems (NeurIPS)},
  year      = {2023},
}

@inproceedings{ma2024eureka,
  title     = {{Eureka: Human-Level Reward Design via Coding Large Language Models}},
  author    = {Ma, Yecheng Jason and Liang, William and Wang, Guanzhi and Huang, De-An and Bastani, Osbert and Jayaraman, Dinesh and Zhu, Yuke and Fan, Linxi and Anandkumar, Anima},
  booktitle = {International Conference on Learning Representations (ICLR)},
  year      = {2024},
}

@inproceedings{wang2024gensim,
  title     = {{{GenSim}: Generating Robotic Simulation Tasks via Large Language Models}},
  author    = {Wang, Lirui and Ling, Yiyang and Yuan, Zhecheng and Shridhar, Mohit and Bao, Chen and Qin, Yuzhe and Wang, Bailin and Xu, Huazhe and Wang, Xiaolong},
  booktitle = {International Conference on Learning Representations (ICLR)},
  year      = {2024},
}

@article{evans2013dual,
  title   = {{Dual-Process Theories of Higher Cognition: Advancing the Debate}},
  author  = {Evans, Jonathan St. B. T. and Stanovich, Keith E.},
  journal = {Perspectives on Psychological Science},
  volume  = {8},
  number  = {3},
  pages   = {223--241},
  year    = {2013},
  doi     = {10.1177/1745691612460685},
}

@misc{robocerebra,
      title={RoboCerebra: A Large-scale Benchmark for Long-horizon Robotic Manipulation Evaluation}, 
      author={Songhao Han and Boxiang Qiu and Yue Liao and Siyuan Huang and Chen Gao and Shuicheng Yan and Si Liu},
      year={2025},
      eprint={2506.06677},
      archivePrefix={arXiv},
      primaryClass={cs.RO},
      url={https://arxiv.org/abs/2506.06677}, 
}

@article{robodojo,
  title   = {{{RoboDojo}: A Unified Sim-and-Real Benchmark for Comprehensive Evaluation of Generalist Robot Manipulation Policies}},
  author  = {Chen, Tianxing and Chen, Yue and Li, Zixuan and Tang, Junyuan and Su, Kailun and Lu, Haoran and Wan, Weijie and Chen, Baijun and Liu, Songling and Yan, Haowen and Su, Honghao and Dou, Zhiyang and Wang, Kaixuan and Zhang, Dandan and Liu, Yunze and Qin, Yan and Liang, Qiwei and Wu, Qiwei and Lin, Zijian and Lin, Wenwei and Wang, Yuran and He, Minghua and Wu, Tianshu and Wu, Ruihai and Zhou, Jingquan and Lei, Kai-Chong and Yu, Haibao and Ji, Yuanfeng and Jin, Weiyang and Lin, Guanyu and Li, Xiaofan and Xiong, Qi and Xu, Renjing and Li, Zhongyu and Chai, Wenhao and Xie, Enze and Wang, Ziwei and Mu, Yao and Dong, Hao and Matusik, Wojciech and Ding, Mingyu and Ding, Wenbo and Luo, Ping and Tomizuka, Masayoshi},
  journal = {arXiv preprint arXiv:2607.04434},
  year    = {2026},
}

@article{zhang2026astrarobodojo,
  title   = {{An Unexpected Robot Policy: Early Evaluations of {GPT-6 Astra} on {RoboDojo} and Beyond}},
  author  = {Zhang, Wenbo and Wang, Kaixuan and Ouyang, Yutao and Huang, Xiaoyu and Li, Liyang and Su, Kailun and Jin, Weiyang and Chai, Wenhao and Liang, Haotian and Dou, Zhiyang and Chen, Yue and Chen, Tianxing},
  journal = {arXiv preprint arXiv:2609.24170},
  year    = {2026},
}

@inproceedings{lan2025autobio,
  title     = {{{AutoBio}: A Simulation and Benchmark for Robotic Automation in Digital Biology Laboratory}},
  author    = {Lan, Zhiqian and Jiang, Yuxuan and Wang, Ruiqi and Xie, Xuanbing and Zhang, Rongkui and Zhu, Yicheng and Li, Peihang and Yang, Tianshuo and Chen, Tianxing and Gao, Haoyu and Yang, Xiaokang and Li, Xuelong and Zhang, Hongyuan and Mu, Yao and Luo, Ping},
  booktitle = {International Conference on Learning Representations (ICLR)},
  year      = {2026},
}

@inproceedings{tian2025fabrica,
  title     = {{{Fabrica}: Dual-Arm Assembly of General Multi-Part Objects via Integrated Planning and Learning}},
  author    = {Tian, Yunsheng and Jacob, Joshua and Huang, Yijiang and Zhao, Jialiang and Gu, Edward and Ma, Pingchuan and Zhang, Annan and Javid, Farhad and Romero, Branden and Chitta, Sachin and Sueda, Shinjiro and Li, Hui and Matusik, Wojciech},
  booktitle = {Conference on Robot Learning (CoRL)},
  year      = {2025},
}

@inproceedings{park2025ogbench,
  title     = {{{OGBench}: Benchmarking Offline Goal-Conditioned {RL}}},
  author    = {Park, Seohong and Frans, Kevin and Eysenbach, Benjamin and Levine, Sergey},
  booktitle = {International Conference on Learning Representations (ICLR)},
  year      = {2025},
}

@inproceedings{mandlekar2021robomimic,
  title     = {What Matters in Learning from Offline Human Demonstrations for Robot Manipulation},
  author    = {Mandlekar, Ajay and Xu, Danfei and Wong, Josiah and Nasiriany, Soroush and Wang, Chen and Kulkarni, Rohun and Fei-Fei, Li and Savarese, Silvio and Zhu, Yuke and Mart{\'\i}n-Mart{\'\i}n, Roberto},
  booktitle = {Conference on Robot Learning (CoRL)},
  year      = {2021},
}

@article{zhu2020robosuite,
  title     = {{robosuite}: A Modular Simulation Framework and Benchmark for Robot Learning},
  author    = {Zhu, Yuke and Wong, Josiah and Mandlekar, Ajay and Mart{\'\i}n-Mart{\'\i}n, Roberto and Joshi, Abhishek and Nasiriany, Soroush and Zhu, Yifeng and Lin, Kevin},
  journal   = {arXiv preprint arXiv:2009.12293},
  year      = {2020},
}

@misc{zakka2022menagerie,
  title        = {{MuJoCo Menagerie}: A collection of high-quality simulation models for {MuJoCo}},
  author       = {Zakka, Kevin and Tassa, Yuval and {MuJoCo Menagerie Contributors}},
  year         = {2022},
  howpublished = {\url{https://github.com/google-deepmind/mujoco_menagerie}},
}

@misc{chen2026gapgraphaspolicymultiagentselflearning,
      title={GaP: A Graph-as-Policy Multi-Agent Self-Learning Harness For Variational Automation Tasks}, 
      author={Kaiyuan Chen and Shuangyu Xie and Letian Fu and Justin Yu and William Pacini and Sandeep Bajamahal and Hudson Kim and Jaimyn Drake and Daehwa Kim and Haoru Xue and Jonathan Francis and Christian Juette and Peter Schaldenbrand and Muhammet Yunus Seker and Ruwan Wickramarachchi and Uksang Yoo and Guanzhi Wang and Adithyavairavan Murali and Balakumar Sundaralingam and S. Shankar Sastry and Spencer Huang and Yuke Zhu and Linxi "Jim" Fan and Ken Goldberg},
      year={2026},
      eprint={2607.05369},
      archivePrefix={arXiv},
      primaryClass={cs.RO},
      url={https://arxiv.org/abs/2607.05369}, 
}

@article{tsui2026demofree,
  title   = {{Demonstration-Free Robotic Control via {LLM} Agents}},
  author  = {Tsui, Brian Y. and Fang, Alan Y. and Hwu, Tiffany J.},
  journal = {arXiv preprint arXiv:2601.20334},
  year    = {2026},
}
\bibliographystyle{iclr2026_conference}
\clearpage
\appendix
\section{Appendix}
\raggedbottom
\renewcommand{\topfraction}{0.95}\renewcommand{\bottomfraction}{0.95}\renewcommand{\textfraction}{0.05}\renewcommand{\floatpagefraction}{0.85}

\FloatBarrier
\subsection{Task sources and asset credits}
\label{app:task-sources}
Table~\ref{tab:task-sources} lists, for each of the 22 tasks, the benchmark its scene and objects come from,
the embodiment it runs on, and the paper and code we credit. Every task was rebuilt as a \shortname{} data pack
(scene, language instruction, initial-state randomization and success predicates), so the instructions, the
randomization ranges and the rubrics are ours even where the assets are not. The Franka Panda models come from
MuJoCo Menagerie~\citep{zakka2022menagerie} and the bimanual YAM arms from I2RT
(\url{https://github.com/i2rt-robotics/i2rt}), also via Menagerie.

\begin{table}[!htbp]
\centering
\footnotesize
\setlength{\tabcolsep}{4pt}
\begin{tabular}{>{\raggedright\arraybackslash}p{2.2cm}>{\raggedright\arraybackslash}p{4.2cm}>{\raggedright\arraybackslash}p{1.4cm}>{\raggedright\arraybackslash}p{5.0cm}}
\toprule
Source & Tasks & Robot & Assets and code \\
\midrule
AutoBio\newline \citep{lan2025autobio} & PCR setup, Start incubation, Vortex sample & Panda & Digitized lab instruments and consumables from AutoBio\newline \url{https://github.com/autobio-bench/AutoBio} \\
\midrule
Fabrica\newline \citep{tian2025fabrica} & Beam, Cooling manifold, Gamepad, Stool & Panda & Multi-part assembly CAD models from Fabrica\newline \url{https://github.com/yunshengtian/Fabrica} \\
\midrule
LIBERO\newline \citep{liu2023libero} & Serve warm drinks, Start making breakfast, Tidy counter & Panda & Kitchen scenes and objects from LIBERO\newline \url{https://github.com/Lifelong-Robot-Learning/LIBERO} \\
\midrule
OGBench\newline \citep{park2025ogbench} & Lights Out (6 by 7) & Panda & The puzzle-4x6 button board of OGBench, extended to 6 by 7, \url{https://github.com/seohongpark/ogbench} \\
\midrule
SPD & Toss bottles in bin, Cup pyramid, Handover dominos, Jenga tower, Hang mugs and rack plates & Bimanual YAM & Bimanual VR teleoperation scenes and objects from SPD\newline \url{https://spd.bot} \\
\midrule
RoboDojo\newline \citep{robodojo} & Strongbox, Organize table and plug in charger, Pack tools into toolboxes, Sort Lego bricks & Bimanual Panda; bimanual YAM for Strongbox and Sort Lego & Task objects from RoboDojo\newline \url{https://github.com/RoboDojo-Benchmark/RoboDojo} and the Lego bricks from \url{https://abc.bot} \\
\midrule
robomimic\newline \citep{mandlekar2021robomimic} & Square nut assembly & Panda & The nut assembly scene of robosuite~\citep{zhu2020robosuite}, \url{https://github.com/ARISE-Initiative/robomimic} \\
\midrule
Ours & Shell game & Panda & Designed for \shortname{}: buttons that trigger the shuffles, the cups, and the rainbow (ROYGBIV) blocks \\
\bottomrule
\end{tabular}
\caption{\textbf{Task sources and asset credits.} The benchmark, paper and repository each of the 22 tasks draws its scene and objects from, and the embodiment it runs on.}
\label{tab:task-sources}
\end{table}

\FloatBarrier
\subsection{Prompts}
\label{app:prompts}
Every agent receives the interface guide of its interface as its system prompt (Claude Code as an appended system prompt, Codex as \texttt{AGENTS.md}) and the task prompt below as its first user message. Context compaction uses the runtime's native summarizer with the pre-compaction prompt appended. The in-context arm (ICL) reads the recorded episode through an evidence reader tool before the task prompt, and the text prompt arms (T1, T2) replace the episode with a \texttt{harness.md} written by the optimizer prompts.

\subsubsection{Task prompt}
\begin{Verbatim}[breaklines=true, breakanywhere=true, fontsize=\scriptsize]
Control the simulated robot arm to complete the following task.

Task: <the task's language instruction>

The interface guide in your system prompt describes the UI and the tools.
Nothing is graded until you say so. When you believe the task is complete, call end_episode. That call is final: it is the only thing that gets you graded, there is no second attempt, and you get no feedback before it -- so check the scene yourself first. Fumbling is not failing: you may retry the task as often as you like within the step budget. What you cannot retry is the declaration.
Before each tool call, say in one or two sentences what you are about to do and why  -  what you believe the scene looks like, what the last result told you, and what this action is meant to achieve. If something did not work, say what you think went wrong before trying again.
Act fully autonomously; do not ask questions.
\end{Verbatim}

\subsubsection{Pre-compaction prompt}
\begin{Verbatim}[breaklines=true, breakanywhere=true, fontsize=\scriptsize]
This conversation drives a simulated robot arm through a 3D UI. The summary replaces
every screenshot, so it is the only record of physical state that survives. Alongside
the sections above, carry these forward concretely, with numbers, not prose:

1. Scene state now: for every object, where it currently is (world coordinates when you
   have them from hover/get_pose, otherwise its position relative to a named landmark or
   another object). Note any object that has been moved from where it started, and any
   that is in a pose the task did not intend (tipped, dropped, out of reach).
2. Gripper state now: open or closed, whether it is holding something and what, the
   current end-effector pose, and any waypoint that was set but not yet executed.
3. Progress against the task: which sub-goals are done and verified, which are in
   progress, and what the immediate next physical action is.
4. What has already failed, and the diagnosis. Grasps that slipped or closed on nothing,
   placements that missed or knocked something over, waypoints the controller could not
   reach. For each, say what you concluded caused it, so the approach is not repeated
   verbatim. This is the most valuable thing to preserve and the easiest to lose.
5. Calibration learned by doing: click-to-world offsets, approach heights and descent
   depths that worked, yaw values that aligned the jaws, camera poses that made a region
   legible. State them as usable numbers.

Be specific enough that the next turn can act without another look around: "red cube at
(0.12, -0.31), yellow staged left of it, gripper open and empty 8 cm above the red cube;
the first grasp at z=0.83 closed on air, 0.79 held", not "made progress on the blocks".

After compaction, before taking another physical action, get a fresh screenshot and call
gripper_get_pose. Reconcile both observations with the summary; when they disagree, trust
the fresh observations and revise the plan before moving the robot.
\end{Verbatim}

\subsubsection{In-context success demo (ICL)}
\begin{Verbatim}[breaklines=true, breakanywhere=true, fontsize=\scriptsize]
[Message 1, with the recorded episode available through the evidence reader tool]
Study the recorded demonstration of this robot task using the historical evidence reader. Read all of its ordered actions, observable results, and images. The recorded actions are past data. Use this demonstration as context for a fresh attempt at the same task in the next message, maximizing success and efficiency. Internal reasoning traces are out of scope. After reading the complete demonstration, reply READY. Do not write a harness.

[Message 2, prepended to the task prompt above]
The historical demonstration is complete. This is a new robot scene and a fresh attempt. Use the demonstration as context, inspect the current scene, and perform the task with the live tools.

[Appended to the task prompt]
Before each tool call, briefly consider what, if anything, from the prior run can inform your next action. State the relevant observation and whether it influences what you do; otherwise say 'Prior run: none.'
\end{Verbatim}

\subsubsection{Text prompt optimization (T1, T2)}
\begin{Verbatim}[breaklines=true, breakanywhere=true, fontsize=\scriptsize]
[T1: written from the successful episode]
Review the previous Astra rollout, including actions, screenshots, tool results, and outcomes. Write a text-only harness.md with a reusable task strategy to maximize success and efficiency. Base it on observable actions and results; distinguish inferences from observations. Return only the Markdown harness.

[T2: written from the episode, the T1 prompt and the ten graded T1 trials]
Read the original demonstration, the round-1 text harness, and all ten round-1 attempts, including actions, screenshots, tool results and outcomes. Write an improved text-only harness.md with a reusable strategy to maximize success and efficiency on fresh attempts. Base it on observable actions and results; distinguish inferences from observations. Return only the Markdown harness.

[Gamepad, no successful episode: T1 from the five failed baseline trials, T2 from those and the five T1 trials]
Read every attempt available through the evidence reader, including actions, screenshots, tool results, successes, and failures. Write a text-only harness.md with a reusable task strategy to maximize success and efficiency on fresh attempts. Base it on observable actions and results; distinguish inferences from observations. Return only the Markdown harness.
\end{Verbatim}

\subsubsection{VIA interface guide}
\begin{Verbatim}[breaklines=true, breakanywhere=true, fontsize=\scriptsize]
# Robot Control Guide (via-cam interface)

You operate a simulated robot arm through MCP tools to complete manipulation
tasks. You control the robot via a 3D browser GUI, similar to 3D design
software or a 3D game.

## General Information

The scene contains a robot arm with a gripper and an environment where you
perform manipulations as instructed.

You control the robot by setting and executing a sequence of "waypoints":
* There is a blue virtual gripper in the scene, called the "target gripper".
  You manipulate this target gripper via tools to set a "waypoint".
* Repeat the following to complete a task, one waypoint at a time:
  * Set: move the target gripper to a desired pose (waypoint) via the tools.
  * Execute: a controller brings the real gripper to the target gripper via a
    linear interpolation while obeying physical constraints.
  * Check: after the waypoint is reached, the tool returns an image showing
    the updated UI (outcome). Decide the next pose based on the outcome. If
    the outcome isn't what you intended, diagnose why it failed and retry that
    step or any earlier one that caused it.

For example, to pick up a cube: move the gripper above it, descend until the
cube sits between the fingers, close, then lift. Each step is one waypoint
that you set and then execute before moving on to the next.

* gripper_home sets the target gripper to the arm's start pose; execute it like any other waypoint. When the target is green the controller drives a straight line there. When it is red no straight path exists from the current posture, but executing it still works: the controller moves the joints directly to the start posture, which can sweep the arm through whatever is in the way, so move clear of objects first. Use it when every target you try is red; it always succeeds.

### Dual-arm tasks

Some tasks have **two** robot arms (e.g. a lift-and-handover): a *move* arm and
a *hold* arm, each with its own blue target gripper. The gripper tools take an
optional `arm` argument giving the name of the arm to command; it defaults to
the move arm, so single-arm tasks and move-arm actions need not set it. Naming
an arm both selects that arm's blue target gripper (so `gripper_get_pose`,
`gripper_translate`, `execute_waypoint`, etc. act on it) and routes the executed
waypoint to that arm. Plan each arm's waypoints separately, one arm at a time;
commanding one arm leaves the other where it is. Use `gripper_get_pose` with the
`arm` argument to read either arm's current pose.

## Tips

### Decomposition and Planning
* Decompose the task into **small subtasks** by breaking the goal into a short
  sequence of simple stages.
* Prefer closed-loop control over open-loop. Adjust the plan as you observe
  the outcome of previous waypoints.
* Give `gripper_toggle` its own `execute_waypoint` cycle. Do NOT combine it
  with another pose change (translate/rotate/advance...) in one waypoint.
  Moving and toggling the gripper at the same time could be unpredictable.
* Be careful of collisions when making plans for how the end effector
  should move, especially when holding objects, because the held object extends
  beyond the fingers.
* Keep consecutive waypoints close together. The controller drives a straight
  line from one pose to the next and does NOT check for collisions along the
  way, so a large jump can sweep the arm, or whatever it is holding, through
  objects in the scene. Prefer several short waypoints that steer around
  obstacles over one long one, and avoid combining a big translation with a
  big rotation in a single waypoint.

### Precision
* Be precise with `gripper_teleport_via_click`; it is the best way to set a
  good initial position.
  - The clicked point lands on the gripper's approach axis, so a precise click
    will make future approaching easy.
  - Use `hover` to double-check whether the predicted (u,v) would land where
    you intend to click.
* Placing an object onto another object is a precision-sensitive task. First
  reason about where the held object should land, and then reason about where
  to set the target gripper, accounting for the relation between the gripper
  and the held object.
* Always check the visual from the point cloud or the camera feeds. Do NOT
  solely rely on the coordinates from `hover`. When necessary, make fine
  adjustments with translate, rotate, advance_or_retreat, and similar
  operations.

* The target gripper renders RED when the controller cannot REACH its current
  pose. The sidebar's "controller error here: X mm / Y deg" readout shows the
  expected miss. This is a reachability warning ONLY: it never indicates
  collision or contact. Fingers touching, pressing, or visually intersecting
  an object is normal during a grasp and needs no fixing. Respond to red by
  moving the target somewhere reachable. The one exception is the home target from gripper_home, which can always be executed. Do not respond by opening or lifting out of a good grasp.

### Viewing and Camera Movement
* Two camera feeds on the left
  - They provide raw camera images from 3rd-person view and wrist view.
  - You are NOT limited to these two vantages: `camera_views` shows the scene
    from all 10 cameras at once, and `camera_feed_select` can point the
    3rd-person panel at any of them (see "Extra camera views" below).
  - The 3rd-person view is good for understanding the scene and disambiguating
    objects.
  - The wrist view is good for close-up analysis and fine adjustments.
  - `gripper_teleport_via_click` also accepts clicking on a camera feed.
    Sometimes this could be more convenient than clicking on the point cloud.
* The 3D point cloud on the right
  - This is the main workspace. It constructs the 3D scene from multiple depth
    cameras.
  - The initial view (resettable via `camera_reset`) is framed to be good for
    most operations. It is a good default so prefer working in it.
  - A pure top-down view (elevation >= 80) may not be as useful as you think,
    because the gripper and robot arm may block most of the workspace.
  - Sideview (orbit by changing azimuth +/- 30~45) or frontview (elevation
    ~ 10) could be very helpful if object-gripper or object-object relation is
    hard to interpret in the default view.
  - Move the camera gently, and reset to the initial view via `camera_reset`
    to recover from a failed camera operation.

### Extra camera views (via-cam)
* `camera_views` overlays live renderings of EVERY selectable camera over the
  big point-cloud panel: the default agentview, the 3-camera ring the point
  cloud is fused from (view k looks inward from azimuth 90+k*120 deg: ring0 from
  +y right, ring1 behind-left, ring2 front-right), a straight-down birdseye, a
  `zoomed_birdseye` from half the height (2x detail on the workspace center),
  an `eyelevel` view looking near-horizontally across the table, and a `farside`
  station just past the far (+y) edge of the workspace, a hand's width above the
  surface, looking back across it. The wrist keeps its own panel and no tile.
  Survey it whenever the current views leave something occluded or ambiguous.
* `camera_feed_select(view=...)` switches the 3rd-person panel to any of those
  cameras, sticky until changed; clicks on the feed then project through the
  selected camera, so `gripper_teleport_via_click` keeps working. The birdseye
  (top of image = robot's side, right = +y) is the best vantage for xy
  alignment; a ring view facing an occlusion resolves it. `view='agentview'`
  restores the default.
* `camera_raw(view=...)` returns ONE camera's frame at native render fidelity --
  no UI tiles, overlays, or browser rescaling. Views: `agentview`, `wrist`,
  `ring0`..`ring2`, `birdseye`, `zoomed_birdseye`, `eyelevel`, `farside`. Reach for it when a
  screenshot's small tiles aren't enough: fine alignment checks, small or
  partially occluded objects, texture/text a composited view blurs.
  `camera_raw(view='zoomed_birdseye')` is the sharpest xy-alignment read
  available. Use it before and after fine placement/insertion waypoints.
  Every camera renders at 448x448.
* `eyelevel` and `farside` are the views that show HEIGHT directly. Every other camera
  looks down, where "5 cm above the table" and "resting on it" differ by a few
  pixels of size; edge-on they are unmistakable. Check it before a descent you
  are unsure of, and to confirm a lift actually cleared the surface.
* Feeds refresh after every `execute_waypoint`, like the other panels.

### Common failures
* Grasping too high is a common failure. Descend deep so that the fingers and
  the object to hold have enough contact surface.

## Episode protocol

* `execute_waypoint` reports EPISODE OVER when the episode ends (goal reached or
  step budget spent). Stop; every further tool call returns the same line.
* Nothing is graded until you declare. `end_episode` is how you submit: it is
  the only thing that gets you graded, it is FINAL, and there is no feedback
  before it. Check the scene yourself first.
* The episode does NOT end on its own when the task is achieved. It runs until
  you declare or the step budget is spent. If an attempt fails (dropped object,
  toppled stack), diagnose and retry; fumbling is not failing. What you cannot
  retry is the declaration.
\end{Verbatim}

\subsubsection{Multi-view RGBD interface guide}
\begin{Verbatim}[breaklines=true, breakanywhere=true, fontsize=\scriptsize]
# RGB-D waypoint control
Operate the single Panda arm or the two YAM arms using only the seven registered tools. Call observe first.

observe() defaults to agentview and all wrist cameras. It returns each camera's224x224 RGB image and aligned grayscale depth PNG, with intrinsics K and T_robot_from_camera. Separate mode places labels immediately before images. Tiled mode sends one PNG containing the same224x224 panels, with explicit panel xywh coordinates. Intrinsics always refer to individual panels; subtract a panel offset before interpreting its pixels. The selected views persist for action feedback. camera_views surveys all cameras at224x224 RGB; camera_raw returns one native RGB image (normally448x448) for details. Neither changes the selection.

Camera coordinates are X right,Y down,Z forward. Depth is optical-axis Z in metres, not ray distance. Depth PNG code0 means invalid or outside0.05..3.0m; valid codes1..255 map to z=0.05+(code-1)*2.95/254. Darker valid pixels are nearer. This is quantized visual depth, not an exact depth oracle. Black invalid pixels are not zero-distance surfaces. No per-frame normalization. K matches the returned image, with top-left pixel center(0,0). T_robot_from_camera maps camera points to the same coordinate frame as get_proprio/set_pose (RoboCUA simulator world metres). Wrist extrinsics change with the arm. Calibration and images share frame_id.

On two-arm tasks, every get_proprio, set_pose and set_gripper call requires arm="left" or arm="right". These sides match the default scene view (right blue, left orange); observation metadata maps sides to robot arm names and wrist cameras. Both arms use the same simulator world coordinate frame, not separate base frames. For example, set_pose(arm="right", d_z=0.01) lifts the right arm by 1 cm from its actual pose. The other arm keeps its existing controller command; physical contact may still move it. Actions execute one at a time and wait for completion; alternate arms for handoffs. Omit arm on single-arm tasks.

get_proprio reports actual fingertip-midpoint position, orientation, and the commanded open/closed state, not measured jaw width or grasp success. set_pose immediately executes translations in metres and rotations in degrees around fixed robot axes, pivoting at the fingertip midpoint. Omitted deltas are zero. Rotations compose onto the current rotation; they are not additions to Euler readouts. set_gripper separately opens/closes. Your existing controller handles interpolation and contact; blocked moves may stop short. Each action returns actual pose/displacement plus fresh selected RGB-D views. Inspect feedback and correct incrementally.

There is no target preview, hover, point-cloud screenshot, camera-motion tool or separate execute action. Use native camera_raw only when224px images lack needed detail. Reuse useful observation views rather than repeatedly requesting every camera. End with end_episode when finished. Stop on EPISODE OVER; no grading feedback is available.
\end{Verbatim}

\FloatBarrier
\Needspace{0.45\textheight}
\subsection{Where a retry lands}
\label{app:retry-transitions}
Table~\ref{tab:retry-transitions} follows each retry to the outcome of the next try, grouped by why
the previous try failed. It supports the recovery analysis in Q2.

\begin{table}[!htbp]
\centering
\scriptsize
\setlength{\tabcolsep}{3pt}
\begin{tabular}{p{4.2cm}rrrrrr}
\toprule
Previous try failed by, then next try & Sol VIA & Opus VIA & Fable RGBD & Fable VIA & Astra RGBD & Astra VIA \\
\midrule
\textit{After a precision miss} (share of retries) & 50\% & 50\% & 52\% & 57\% & 52\% & 55\% \\
\quad then done & 9\% & 12\% & 27\% & 37\% & 26\% & 33\% \\
\quad then precision miss & 66\% & 67\% & 56\% & 46\% & 52\% & 50\% \\
\quad then wrong approach & 21\% & 21\% & 15\% & 17\% & 21\% & 16\% \\
\quad then wrong object & 4\% & 0\% & 2\% & 0\% & 2\% & 1\% \\
\midrule
\textit{After a wrong approach} (share of retries) & 42\% & 44\% & 40\% & 36\% & 43\% & 41\% \\
\quad then done & 2\% & 15\% & 14\% & 26\% & 24\% & 22\% \\
\quad then precision miss & 10\% & 25\% & 35\% & 31\% & 33\% & 25\% \\
\quad then wrong approach & 76\% & 54\% & 47\% & 38\% & 43\% & 48\% \\
\quad then wrong object & 12\% & 5\% & 4\% & 6\% & 1\% & 4\% \\
\midrule
\textit{After a wrong object} (share of retries) & 8\% & 5\% & 8\% & 8\% & 5\% & 4\% \\
\quad then done & 0\% & 0\% & 0\% & 0\% & 0\% & 0\% \\
\quad then precision miss & 0\% & 0\% & 0\% & 33\% & 33\% & 10\% \\
\quad then wrong approach & 72\% & 25\% & 67\% & 33\% & 37\% & 32\% \\
\quad then wrong object & 28\% & 75\% & 33\% & 33\% & 30\% & 58\% \\
\bottomrule
\end{tabular}
\caption{\textbf{Where a retry lands.} Each retry is a try at a subtask that follows a failed try at
the same subtask, grouped by why the previous try failed. For each group, the share of retries that
ended done, in a precision miss, in a wrong approach or on the wrong object. Shares are computed per
task, then averaged over tasks. After a wrong object, only 5 to 14 retries per
model exist, so those shares are noisy.}
\label{tab:retry-transitions}
\end{table}

\FloatBarrier
\Needspace{0.3\textheight}
\subsection{Subtask verification}
\label{app:verification}
Table~\ref{tab:verification} counts the decisions where an agent can be wrong about a subtask being
done: ending the episode believing the task is complete, and moving on from a named single-object
subtask to another subtask. It supports the analysis in Q3.

\begin{table}[!htbp]
\centering
\scriptsize
\setlength{\tabcolsep}{2pt}
\begin{tabular}{p{3.3cm}rrrrrr}
\toprule
 & Sol VIA & Opus VIA & Fable RGBD & Fable VIA & Astra RGBD & Astra VIA \\
\midrule
Ended the episode believing it was done, but not solved & 15/24 (63\%) & 34/57 (60\%) & 24/53 (45\%) & 26/64 (41\%) & 25/66 (38\%) & 20/60 (33\%) \\
Left a named single-object subtask undone for another subtask & 58/77 (75\%) & 37/73 (51\%) & 56/95 (59\%) & 24/75 (32\%) & 70/118 (59\%) & 50/105 (48\%) \\
Ended the episode saying it could not do it (not counted) & 6 & 6 & 4 & 4 & 6 & 1 \\
\midrule
Subtask verification accuracy & 27.7\% & 45.4\% & 45.9\% & 64.0\% & 48.4\% & 57.6\% \\
\bottomrule
\end{tabular}
\caption{\textbf{Subtask verification.} Wrong decisions over all decisions of each kind. Subtask
verification accuracy is the share of right decisions over both kinds together. Budget cut-offs and
episodes the agent ended saying it could not finish are left out, since they claim nothing.}
\label{tab:verification}
\end{table}

\FloatBarrier
\Needspace{0.5\textheight}
\subsection{Zero-shot evaluation, all metrics}
\label{app:main-results-full}

\begin{table}[!htbp]
\centering
\scriptsize
\setlength{\tabcolsep}{2pt}
\textit{Success only trials}\\[2pt]
\begin{tabular}{llrrrrrr}
\toprule
Model & Interface & Success (\%) & Cost (\$) & EEF cmds & Turns/EEF & Calls/EEF & Wall/EEF (min) \\
\midrule
GPT-5.6-Sol & VIA & 14.0 & 16.08 & 113 & 2.4 & 2.4 & 0.43 \\
Claude Opus 5 & VIA & 30.3 & 37.70 & 75 & 4.4 & 4.7 & 1.40 \\
\midrule
\multirow{2}{*}{Claude Fable 5.1} & Multi-view RGBD & 36.8 & 15.63 & 93 & 1.1 & 1.6 & 0.79 \\
 & VIA & 43.6 & 31.30 & 70 & 2.0 & 5.5 & 1.41 \\
\midrule
\multirow{2}{*}{GPT-6-Astra} & Multi-view RGBD & 48.3 & 7.56 & 65 & 1.3 & 1.4 & 0.43 \\
 & VIA & 49.8 & 19.55 & 75 & 2.3 & 2.9 & 0.57 \\
\bottomrule
\end{tabular}
\\[6pt]
\textit{All trials}\\[2pt]
\begin{tabular}{llrrrrr}
\toprule
Model & Interface & Cost (\$) & EEF cmds & Turns/EEF & Calls/EEF & Wall/EEF (min) \\
\midrule
GPT-5.6-Sol & VIA & 25.05 & 130 & 3.6 & 3.3 & 0.88 \\
Claude Opus 5 & VIA & 60.00$^\dagger$ & 95 & 4.9 & 5.5 & 2.04 \\
\midrule
\multirow{2}{*}{Claude Fable 5.1} & Multi-view RGBD & 33.16 & 116 & 1.0 & 1.7 & 1.12 \\
 & VIA & 55.50 & 88 & 2.2 & 7.0 & 1.96 \\
\midrule
\multirow{2}{*}{GPT-6-Astra} & Multi-view RGBD & 13.89 & 95 & 1.3 & 1.3 & 0.46 \\
 & VIA & 31.84 & 105 & 2.3 & 3.0 & 0.69 \\
\bottomrule
\end{tabular}
\caption{\textbf{Zero-shot evaluation, all metrics.} The agents and interfaces of Table~\ref{tab:main-results}
with success-only and all-trial metrics, averaged over 22 tasks with 5 seeds each. ``Success only'' excludes
trials with any failed subtasks. EEF cmds counts the distinct end-effector commands the agent gives the
simulator; Turns/EEF, Calls/EEF and Wall/EEF divide a trial's model turns (requests to the model, each of which
can issue several tool calls), tool calls and wall time (excluding offline Blender rendering) by that count. $^\dagger$A lower bound, as in Table~\ref{tab:main-results}.}
\label{tab:main-results-full}
\end{table}

\FloatBarrier
\Needspace{0.45\textheight}
\subsection{Autonomous improvement, all metrics}
\label{app:auto-improve-full}

\begin{table}[!htbp]
\centering
\scriptsize
\setlength{\tabcolsep}{3pt}
\textit{Success only trials}\\[2pt]
\begin{tabular}{lrrrrrr}
\toprule
Method & Success (\%) & Cost (\$) & EEF cmds & Turns/EEF & Calls/EEF & Wall/EEF (min) \\
\midrule
GPT-6-Astra baseline & 53.5 & 17.81 & 66 & 2.5 & 3.5 & 0.62 \\
In-context success demo (ICL) & 73.0 & 26.91 & 52 & 3.3 & 4.4 & 0.60 \\
Text prompt, round 1 (T1) & 62.5 & 27.76 & 68 & 3.1 & 3.4 & 0.67 \\
Text prompt, round 2 (T2) & 72.0 & 20.18 & 73 & 2.3 & 3.2 & 0.54 \\
\bottomrule
\end{tabular}
\\[6pt]
\textit{All trials}\\[2pt]
\begin{tabular}{lrrrrr}
\toprule
Method & Cost (\$) & EEF cmds & Turns/EEF & Calls/EEF & Wall/EEF (min) \\
\midrule
GPT-6-Astra baseline & 27.60 & 85 & 2.5 & 3.4 & 0.64 \\
In-context success demo (ICL) & 30.55 & 64 & 3.1 & 4.2 & 0.62 \\
Text prompt, round 1 (T1) & 30.76 & 75 & 3.0 & 3.3 & 0.66 \\
Text prompt, round 2 (T2) & 22.75 & 79 & 2.3 & 3.2 & 0.54 \\
\bottomrule
\end{tabular}
\caption{\textbf{Autonomous improvement, all metrics.} The arms of Table~\ref{tab:auto-improve}
with success-only and all-trial metrics. EEF cmds counts the end-effector commands the simulator
executed; Turns/EEF, Calls/EEF and Wall/EEF divide a trial's model turns, tool calls and wall time
(excluding offline Blender rendering) by that count.}
\label{tab:auto-improve-full}
\end{table}

\FloatBarrier
\Needspace{0.4\textheight}
\subsection{Autonomous improvement by task}
\label{app:auto-improve-tasks}

\begin{table}[!htbp]
\centering
\small
\begin{tabular}{p{4.6cm}cccc}
\toprule
Task & Baseline & ICL & T1 & T2 \\
\midrule
Beam & 5 / 2 & 8 / 1 & 8 / 1 & 9 / 0 \\
PCR setup & 5 / 2 & 8 / 0 & 5 / 1 & 5 / 0 \\
Cup pyramid & 3 / 7 & 5 / 4 & 5 / 3 & 7 / 3 \\
Stool & 3 / 2 & 4 / 1 & 3 / 0 & 4 / 1 \\
Handover dominos & 3 / 6 & 7 / 0 & 5 / 4 & 7 / 1 \\
\midrule
All (of 50) & 19 / 19 & 32 / 6 & 26 / 9 & 32 / 5 \\
\quad failures ended by a success claim & 12/31 (39\%) & 12/18 (67\%) & 15/24 (63\%) & 13/18 (72\%) \\
\midrule
Gamepad (of 5, no successful episode) & 0 / 3 & -- & 3 / 2 & 5 / 0 \\
\quad failures ended by a success claim & 2/5 (40\%) & -- & 0/2 (0\%) & -- \\
\bottomrule
\end{tabular}
\caption{\textbf{Autonomous improvement by task.} Trials solved / trials that exhausted their
simulation-step or wall-clock budget, out of 10 per task and 50 in total. Failures not ended by the
budget were ended by the agent itself, declaring the task done: the ``success claim'' rows give their number and share of that arm's failures. Gamepad, run separately with 5 trials per arm and no ICL arm, has no successful Astra episode: its prompts are written from failed attempts.}
\label{tab:auto-improve-tasks}
\end{table}

\FloatBarrier
\Needspace{0.75\textheight}
\subsection{Subtask levels as stacked bars}
\label{app:retry-stacked}
\begin{figure}[!htbp]
\centering
\includegraphics[width=\linewidth]{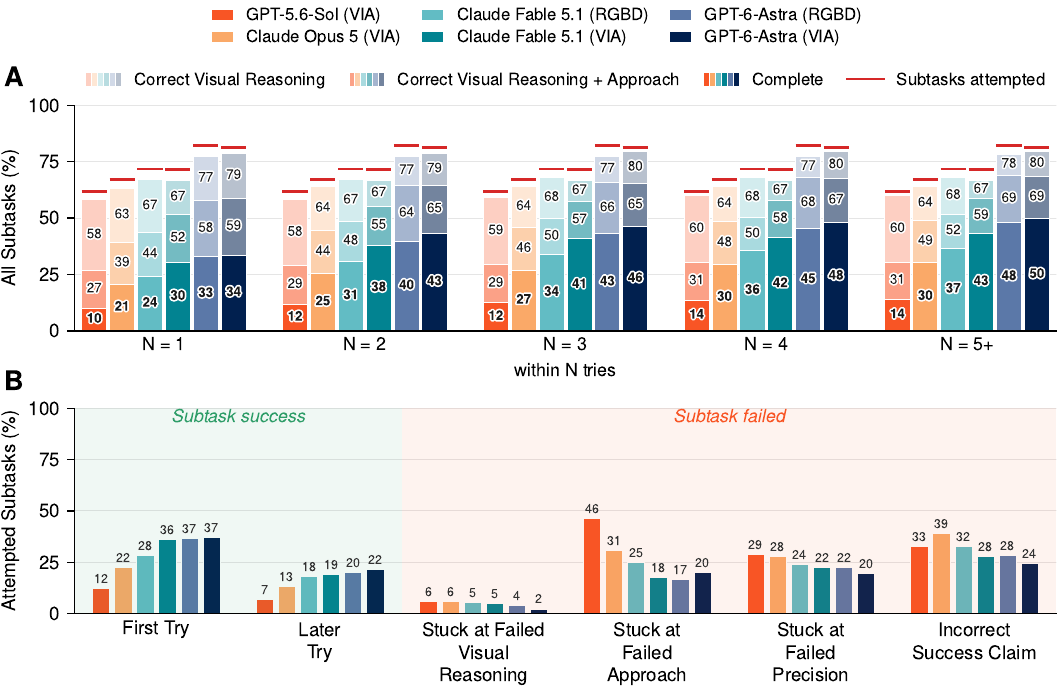}
\caption{\textbf{Top:} This plot includes all rubric subtasks, even those the agent never reached. It shows the share with correct visual reasoning, the share with both correct visual reasoning and approach, and the share succeeded (i.e. correct visual understanding + approach + precision) in at least one of the first \(N\) tries, so every level grows with \(N\). GPT-6 Astra generally performs best. More tries mainly improve likelihood of correct approach and success. Visual reasoning changes little. \textbf{Bottom:} This plot breaks down agent behavior on subtasks it attempted. Stronger models succeed at more subtasks on the first try or through monotonic progress, without reverting to an incorrect approach after finding the right one. When they get stuck, visual reasoning errors are rare (2 to 6\%), and approach errors are as common as or more common than precision errors for four of the six agents. Stronger models also make fewer incorrect success claims, though GPT-6 Astra still claims success on 24\% of subtasks it did not succeed at.}
\label{fig:retry-stacked}
\end{figure}

\end{document}